Włodzisław Duch
Department of Informatics, and Neurocognitive Laboratory
Nicolaus Copernicus University
https://orcid.org/0000-0001-7882-4729

# Artificial intelligence and the limits of the humanities.

The complexity of cultures in the modern world is now beyond human comprehension. Cognitive sciences cast doubts on the traditional explanations based on mental models. The core subjects in humanities may lose their importance. Humanities have to adapt to the digital age. New, interdisciplinary branches of humanities emerge. Instant access to information will be replaced by instant access to knowledge. Understanding the cognitive limitations of humans and the opportunities opened by the development of artificial intelligence and interdisciplinary research necessary to address global challenges is the key to the revitalization of humanities. Artificial intelligence will radically change humanities, from art to political sciences and philosophy, making these disciplines attractive to students and enabling them to go beyond current limitations.

*Nothing is more terrible than to see
ignorance in action.
— Johann Wolfgang von Goethe,
Maxims and Reflections (1826)*

## Introduction: three worlds.

For most of human history, we have seen ignorance in action. Science explains phenomena and increases our understanding of physical and mental processes, but our limited cognitive abilities may impose fundamental limitations on what we can understand. Science should help to create a better world, not only through technology that improves our material situation. "Better" means more satisfying, deepening our understanding of the three worlds: physical, mental, and cultural[1]. The material world is the foundation. All matter is a configuration of atoms, creating chemical bonds and complex structures. Atoms interact with each other, creating molecules. After billions of years, this led to the creation of biological organisms with brains of incredible complexity capable of creating mental states. This inner world reflects selected aspects of the physical world[2]. Interactions of neural activations, structured by experience, led to a unique, subjective, complex mental world that increases the chances of survival and maintains homeostasis. Physics, chemistry, biology, medicine, psychology, and cognitive sciences are among many branches of science that try to discover processes that the physical and mental worlds are based upon. Some of these discoveries are translated into technologies, including medical and environmental technologies that help to maintain our physical and mental health.

Cognitive science is an interdisciplinary attempt to explain the nature of the mind based on brain processes. Brain research reveals nature's solutions in responding to environmental challenges. However, a complete understanding of the brain is insufficient to fully understand the mental world. Explaining the reasons for the emergence of specific mental states, emotions, aesthetic appreciations, and understanding the "third world" of culture requires detailed knowledge of the individual history of a person since her/his birth, a collective history of the local culture, and even the evolutionary history of our species. Disciplines referring to the human experience, such as cultural anthropology, philosophy, literature, history, language, religion, art, and music, study this world. Such studies are based on analytic, critical, and frequently speculative methods. The main question addressed in this article is: what are the limitations of such approaches, and can they be overcome using tools created by modern science? In particular, recent programs in digital humanities and developments in artificial intelligence large language models empower us to address problems beyond human comprehension.

---

[1] Karl Popper has introduced this popular division of realms of existence, see also M. Heller, *Philosophy in Science: An Historical Introduction* (Springer, 2011).
[2] Fascinating story of the origin of life and development of brains is in: LeDoux, Joseph. *The Deep History of Ourselves: The Four-Billion-Year Story of How We Got Conscious Brains*. New York City: Viking, 2019.

An evolutionary perspective allows for a partial answer to the question of "why we are the way we are." E.O. Wilson's "Sociobiology, The New Synthesis,"[3] published in 1975, was denounced fervently as an "ideology" that justifies social inequalities for decades. After 25 years, behaviorist John Alcock wrote "The Triumph of Sociobiology," summarizing achievements of sociobiology that provide the best explanation of social behavior achieved so far. The high complexity of the mammal brain requires specialization of brain areas for perception, attention, motor, memory, spatial orientation, planning, executive control, and other functions. Chances of survival were increased by intelligence distributed in social structures, with specialized but closely cooperating brains. In the case of insects living in colonies, interactions between separate brains are mediated through chemical senses (smell, taste), tactile (vibrations), and visual perception. Fish swimming in large schools or flying birds' flocks increase their chances of survival. With much bigger brains than insects, they can survive longer on their own. Insects need much stronger cooperation; their individual life is not that important. Local neural ganglia control eight limbs of an octopus, allowing them to perform semi-autonomous tasks, with the central brain coordinating sensory, motor, and associative information processing. Humans have much bigger brains with strongly integrated internal communication between specialized brain areas, facilitating better communication within and between ourselves. Communication within the brain enables understanding and planning activities at the mental level. Increased cooperation led to the creation of large social structures that have reached a much higher level of competence in solving basic existential problems (food, shelter, safety). The formation of complex societies led to entirely new problems (wars, famine, pandemics) that required even more cooperation and accumulation of knowledge to solve them. Social sciences have endeavored to devise scientific techniques to comprehend social phenomena in a way that can be generalized. However, only in recent times cognitive social neuroscience has linked social sciences with biological underpinnings.

---

[3] Wilson, E. O. (1975). Sociobiology: The New Synthesis. Belknap Press.

Biologists consider as culture any behavior that is learned by groups. More precisely, "Culture is socially learned information stored in individuals' brains that is capable of affecting behavior."[4] Competing societies develop different social structures and cultures. This process has been observed in social animals, especially in non-human primates[5]. The "cultural primatology"[6] field has documented differences in over 40 behavioral patterns among chimpanzee communities, including grooming, courtship, and tool usage. Language and accumulation of knowledge allowed human cultures to reach an extremely high level of sophistication and develop local cultures, belief systems, and science. Information and communication technologies allowed for more cooperation, leading to globalization.

In the next section, some bridges between humanities and various branches of science are discussed. We take a closer look at challenges to humanities in the age of artificial intelligence, present some capabilities of generative AI and large language models, discuss cognitive limitations of our brains, briefly summarize the current situation of the humanities, and finish with a description of opportunities opening in digital humanities and applications of AI.

## Bridges to humanities

Cultures of closed societies or subgroups within society are relatively stable and well-defined. Population growth, interaction between different societies, and the development of communication technology led to the emergence of many different subcultures. People today identify with an increasing variety of artistic movements, musical styles, fashions, political philosophies, religious sects, new age groups, gender communities, and beliefs. Some of these subcultures are ephemeral, and some persist for a long time; their popularity grows and wanes. Understanding why and how this happens is an important challenge. In 1976, Richard Dawkins introduced memetics[7] as the science of cultural information transfer, trying to understand why new information is sometimes quickly spread and becomes a meme. After a few years of general interest, memetics became largely abandoned[8]. Without the physical substrate of memes, it was disconnected from other science fields and replaced by ideas based on gene-culture co-evolution or dual inheritance theory. Relations between genes, epigenetics, environment, social structure, and culture that new theories describe are fascinating (some examples are presented in the book by Sapolsky[9]). Memetics was concerned with much faster, short-lived processes. Memes can be understood as memory states that are easily created and maintained in the conceptual networks prevailing in a given subculture[10]. The failure of memetics as a theory of cultural evolution shows the importance of finding bridges between different branches of science.

---

[7] Richard Dawkins, *The Selfish Gene: 30th Anniversary Edition* (OUP Oxford, 2006).
[8] Radim Chvaja, 'Why Did Memetics Fail? Comparative Case Study', *Perspectives on Science*, 28.4 (2020), 542–570.

We can imagine levels of existence ordered by the supervenience relation: differences at the lower level are necessary for differences at the higher level. Starting with the material world, mental, social, and cultural levels supervene on each other, each emerging from the lower, simpler level. Science is the basis for the development of technology that improves material living conditions, including physical and mental health. Social sciences study relationships between individuals and the formation of whole societies. Finally, within societies, different cultures and subcultures emerge. Humanities study extremely complex phenomena, trying to understand many aspects of culture, including the influence of physical and mental domains on history, art, language, literature, law, politics, philosophy, and religion. By studying the world of human creations, humanities reveal the range of possible mental states and human behaviors. Ultimately, this helps us understand human nature and points science towards creating a world centered around human needs. Such studies should help to alleviate "the psychological misery of mankind" (Freud, 1930)[11].

Each branch of science focuses on a selected group of phenomena, developing specialized language and methods. Many bridges exist between different branches of science, but our understanding of human nature is still far from satisfactory. As a result of rapid growth and compartmentalization of knowledge, we know more and more about less and less. A small group of highly educated individuals could keep up with all significant knowledge during the Renaissance. In the early XIX century Johann Wolfgang von Goethe (1749–1832), known as a poet, novelist, and dramatist, wrote:

> As to what I have done as a poet,... I take no pride in it... But that in my century I am the only person who knows the truth in the difficult science of colors—of that, I say, I am not a little proud, and here I have a consciousness of a superiority to many.

---

[9] Robert M. Sapolsky, *Behave: The Biology of Humans at Our Best and Worst* (New York, New York: Penguin Press, 2017).

[10] Włodzisław Duch, 'Memetics and Neural Models of Conspiracy Theories', *Patterns*, 2.11 (2021), 100353.

[11] Freud, Sigmund. "Civilization and Its Discontents". German orig. 1930, Translated by James Strachey. New York: W. W. Norton & Company, 2010.

— Johann Eckermann, Conversations with Goethe[12]

Goethe was indeed a great naturalist who studied optics, anatomy, geology (he collected almost 18000 rock samples), biology, physics, and meteorology (popularizing "Goethe barometer"). Great thinkers of the 18th and 19th centuries, remembered now as poets and philosophers, also had a keen interest in natural sciences. The rapid accumulation of knowledge led to the separation of the branches of science, each with its specialized language and research methods. It created a deep division among people studying different phenomena. School curricula and incompetent teachers have made these divisions even more profound. Our brains did not evolve to handle detailed knowledge about the world. We have reached the limits of what we can learn. Education needs deep reflection, providing a shared knowledge foundation to understand the physical, mental, and cultural worlds. Natural sciences are the basis on which material civilization is built. Social sciences should teach us how to achieve a state of well-being. Humanities help to understand different cultures, history, and ways of perceiving social relations, which is essential in the global world. Technology may guide us in the complex world we live in.

---

[12] Eckermann Johann Peter, *Conversations with Goethe in the Last Years of His Life*, trans. S.M. Fuller (Boston and Cambridge: James Munroe and Company, 1852).

Although the history of humanities can be traced back to antiquity, Google n-gram viewer[13] shows the rise of the term "humanities" in books starting from about 1940, reaching its peak of popularity in the mid-1960s. Since then, the core subjects in humanities have been losing popularity, but hybrid fields are emerging and are here to stay. Writing about new humanities, Jeffrey J. Williams mentioned in his essay "digital humanities, environmental humanities, energy humanities, global humanities, urban humanities, food humanities, medical humanities, legal humanities, and public humanities"[14]. They represent the next stage of adaptation to changes in the world. Science is trying to find answers to big challenges, and human factors are essential to such efforts. Natural sciences contribute new research methods that revolutionize some fields of humanities study, making scientists interested in how their methods work in the real world. Dating methods created by physicists and chemists have been used for a long time in the history and restoration of arts. Archeology uses dating methods and has benefitted dramatically from lidar and satellite imagery observation. Neuroarthistory[15] tries to understand the development of arts and music from the perspective of neural mechanisms behind perception influenced by cultural context. Cognitive history strives to understand how people thought, how their worldview and cognitive abilities evolved, and how it has influenced their decisions[16].

---

[13] https://books.google.com/ngrams/
[14] Jeffrey J. Williams, "The New Humanities. Once-robust fields are being broken up and stripped for parts". In: Endgame. Chronicle Review (The Chronicle of Higher Education, 2020), pp. 25-28.
[15] Onians John, *Neuroarthistory: From Aristotle and Pliny to Baxandall and Zeki* (New Haven Conn. ; London: Yale University Press, 2008).
[16] Dunér David and Ahlberger Christer, *Cognitive History: Mind, Space, and Time*, *Cognitive History* (De Gruyter Oldenbourg, 2019).

Natural language processing (NLP) trained on ancient texts helps to guess missing characters, words, and phrases in damaged cuneiform texts. Automatic translation of texts written in scripts that only a few experts can understand (or have not been yet decoded) is now done on a large scale. In May 2023 large-scale projects to preserve the world's language diversity through massively multilingual text/speech AI models could convert speech to text in over 1100 languages and identify more than 4,000 spoken languages. New language tools give unprecedented power to scholars studying less-known cultures. Education benefits from individual tutoring, from learning a conversational foreign language to programming.

New fields, such as neurophilosophy [17] and neurophenomenology [18], have developed under the pressure of cognitive neurosciences. Embodied cognition become one of the most exciting trends bridging science, psychology, and humanities [19]. Philosophy of mathematics went beyond centuries of discussions between Platonism and constructivism[20], thanks to the analysis of mathematical ideas treated as conceptual metaphors by cognitive linguistics. Neurolinguistics has made significant progress since Sidney Lamb defined it as the basis of language[21] and found practical applications in the development of AI algorithms[22]. Understanding brain processes involved in language comprehension leads to "brain-based semantics," a new representation of concepts that connects the meaning of concepts to specific brain areas activated in different cognitive states[23].

Humanities can provide interesting questions, pointing to gaps in our knowledge about some period of history, the development of civilizations, and reasons for historical changes. In recent years, historians have started to consider the influence of climate changes and great natural catastrophes on ancient societies. Floods, volcanic eruptions, asteroid impacts, earthquakes, and pandemics had a major influence on humanity. New technologies and analysis of economic data show new light on political sentiments and social development. These are just a few selected examples of how humanities have fused with other branches of science.

## Challenges to Humanities in the Age of Artificial Intelligence

Science has yet to address many great challenges, both at the fundamental level and in understanding complex interacting systems. Cosmology has a standard model that is full of mysteries: what caused the sudden expansion of the Universe? Such concepts as inflation field, dark energy, and dark matter are just fancy names that cover our ignorance. On the other hand, understanding ecosystems, climate, cells and biological organisms, culture, and human beliefs may not require any fundamental new knowledge. Complexity emerges from interactions and thus is hard to understand[24]. No simple laws or sets of rules are sufficient to explain the structure and dynamics of complex systems. Traditional analysis of social processes and culture cannot account for their complexity.

---

[23] Jeffrey R. Binder et al. 'Toward a Brain-Based Componential Semantic Representation', *Cognitive Neuropsychology*, 33(3–4) (2016), 130–74.
[24] Murray Gell-Man, who got the Nobel prize for theory of quarks, wrote a book *The Quark and the Jaguar: Adventures in the Simple and the Complex* (New York: St. Martin's Griffin, 1995) stressing these two sides of physics.

Humanities are diverse, from literature, arts, political science, religion, and philosophy to anthropology. Some are close to science, others maintain their distinctiveness. The impact of digital humanities and artificial intelligence will soon be obvious in all these areas. IBM Watson debater has won in live performances during debates with human experts[25]. We listen to the comments of social and political scientists about the current world situation, but health experts analyze data and draw conclusions based on software systems that perform simulations. The spread of public opinions can also be simulated using multi-agent systems.

The experimental non-narrative film "Koyaanisqatsi: Life Out of Balance" was released in 1982, with music composed by Philip Glass. The movie shows rapid changes in the natural and human environment. In the past, cultures had time to adjust to slow changes, and interactions between cultures were sparse. People had little idea of the far-reaching consequences of their actions. Plans for the future were based on expansion, conquering new lands, and enslaving more people. These days, with billions of people on the planet, global interactions are more intricate, and changes spread swiftly and in unexpected ways. Cultures become fluid and are in constant transition. Can we still hope that our intelligence is sufficient to understand a world of such complexity? We can read and learn only a tiny fragment of all relevant sources on any subject[26]. Artificial intelligence has no limitations and can internalize and use all available information in reasoning. In the historic match in Go, the most complex traditional game invented by humans, Google AlphaGo, playing against Korean champion Lee Sedol, made a beautiful move that no human ever would make. After thousands of years of playing Go, humans have learned something new from a creative AI program. In 2019 Lee Sedol retired from playing professional Go, acknowledging that AI is "an entity that cannot be defeated"[27]. Since then, AI superiority has been demonstrated practically in all types of games, including games of chance, such as poker or bridge, that require understanding human intentions. Complex reasoning is better left to the AI systems.

---

[25] Bar-Haim, Roy, Yoav Kantor, Elad Venezian, Yoav Katz, and Noam Slonim, 'Project Debater APIs: Decomposing the AI Grand Challenge' (arXiv, 2021). https://doi.org/10.48550/arXiv.2110.01029
[26] Visualization of human knowledge shows a high-level view of this complexity, cf. Veslava Osinska and Grzegorz Osinski, *Information Visualization Techniques in the Social Sciences and Humanities* (IGI Global, 2018).

Theories in natural sciences are based on empirical data collected in many laboratories, verified in different experiments, and eventually applied in practice. The humanities study complex phenomena, building conceptual models that lack the data to verify them. We do not know what the motivation of individuals in the past was, but cognitive history tries to understand how ancient cultures viewed the world. One of the early attempts to understand the history of science "from the inside," taking into account the milieu of the Copernicus, Kepler, and Newton times, was made by Arthur Koestler[28]. There are still only a few works of this type. Cognitive history is a new field focused on understanding how and what people thought in the past[29].

I have deliberately cited early books and papers that initiated the new trends on the border of sciences and humanities to show that some of these attempts are not new; they started decades ago. Their impact on mainstream research in humanities is accelerating. Understanding the roots of cognitive inertia in the development of all branches of science is an important challenge. Max Planck has made a famous observation about the strong resistance to new ideas: "A new scientific truth does not triumph by convincing its opponents and making them see the light, but rather because its opponents eventually die, and a new generation grows up that is familiar with it."[30]. Most people resist changes and refuse to learn new things outside their narrow domain of expertise. While distorted beliefs, including conspiracy theories, are described in popular books and studied by humanities experts[31], deeper mechanisms that make some brains susceptible to such beliefs have rarely been studied[32].

---

[27] https://en.wikipedia.org/wiki/Lee_Sedol
[28] Arthur Koestler and Herbert Butterfield, *The Sleepwalkers: A History of Man's Changing Vision of the Universe* (Penguin Books, 1990).
[29] Dunér, D., & Ahlberger, C. (2019). Cognitive History: Mind, Space, and Time (D. / A. Christer Dunér, Ed.; 1st edition). De Gruyter Oldenbourg.

Despite great progress in the methodological development of hermeneutics, an attempt to define "objective hermeneutics" in the 1970s, and numerous applications in diverse research fields[33], the hope to gain undisputed knowledge in this way has not been justified. Searching for latent meaning brings objective hermeneutics methods close to natural language processing (NLP), a branch of artificial intelligence. Thanks to the Internet, we have gained instant access to information, retrieving articles from encyclopedias, books, and research papers. We still need to find what to search for and then how to analyze the results in a broader context to find the answers. We are already asking AI systems to find and summarize information and will soon discuss our problems with smartphones. Huge language models that started with GPT-3, now have over a trillion parameters and can answer questions about complex texts in hundreds of languages more accurately than people. They can sustain meaningful dialog on any subject, write articles, and computer code[34]. The fact that AI systems can understand sophisticated questions, analyze images, and summarize relevant knowledge is "more than a little terrifying."[35] Important discoveries in such complex domains as computer algorithms, bioinformatics, chemistry, and material science[36] have already been reported. Open Research Europe has already started a collection of "Artificial Intelligence and the Social Sciences and Humanities"[37] articles.

---

[30] Max Planck (2014). Scientific Autobiography: And Other Papers. Philosophical Library/Open Road.
[31] Michael Shermer, 'Why People Believe Conspiracy Theories', *Skeptic*, 25 (2020), 12–18.
[32] Włodzisław Duch, 'Memetics and Neural Models of Conspiracy Theories', *Patterns*, 2(11) (2021), 100353.
[33] Andreas Wernet, *Hermeneutics and Objective Hermeneutics*. Chap. 16, in Flick, Uwe. The SAGE Handbook of Qualitative Data Analysis. SAGE, 2013.
[34] Remarks of 9 philosophers on capabilities of the GPT-3 model are at this address: https://dailynous.com/2020/07/30/philosophers-gpt-3/

## Generative AI and the near future

In the next few years, instant access to knowledge may become as ubiquitous as access to information is now. Instead of searching, analyzing, and connecting bits and pieces of information, we will ask AI programs for precise answers, arguments, and explanations. We can already talk to the books using ChatPDF. The role of textbooks will change. Hundreds of companies specialize in applications of AI in law. AI is changing legal professions, writing and reviewing contracts, bank transactions, and precedence. Artificial auditors search documents for hidden biases that people frequently miss. Available tools will make traditional education in law schools obsolete. "The disruptions from AI's rapid development are no longer in the distant future. They have arrived …"[38].

---

[35] Farhad Manjoo, "How Do You Know a Human Wrote This?". The New York Times (July 29, 2020).
[36] The AI field is changing too quickly to cite publications. A good source of recent news on all aspects of AI may be found at this page: https://flipboard.com/@wlodzislaw/ai-ci-ml-q2dhj0nuy.
[37] https://open-research-europe.ec.europa.eu/collections/ai-in-ssh/about
[38] Generative AI in the Legal Profession. Special issue of The Practice, Center of the Legal Profession, Harvard Law School. March/April 2023.

Words invoke images in our minds. Artificial intelligence systems can analyze images and videos, describe images from the camera, recognize people and places in photographs, and help blind people. Recent AI software, such as Dall-e by OpenAI, Imagen by Google, or Midjourney, prompted by text requests, create many ingenious variants of images that do not resemble anything that humans have ever created. For example, asking the Dall-E3 program[39] to paint a "cat riding on a scooter with Napoleon's hat, smoking a pipe, in Monet style" creates several images, like the one presented here. After the first generative AI program associating descriptions with images has been shown, ten different systems have been developed in a few months of the summer of 2022. Immediately, artists, illustrators, and designers started to use them (YouTube contains many videos illustrating the capabilities of such programs). AI programs can create images according to personal aesthetic preferences. Programs trained with natural photographs can develop a unique sense of aesthetics without training on paintings created by humans. AI-generated art has already won art competitions. Writing text prompts to generate interesting images became a new, highly desired skill. Humans are becoming art curators, leaving the technical side of drawing or painting techniques to AI software.

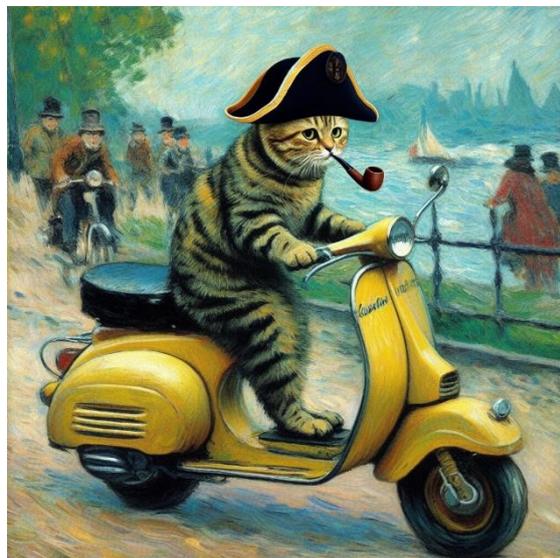

Text-to-image programs have been trained on billions of photos, paintings, and sculptures. Their imagination and ability to blend deep representations that are expressed in visual form are much richer than any human can ever achieve. Why should the imagery of text-to-image software exceed human abilities? Our visual system is very complex, consisting of dozens of specialized brain areas. Countless images seen in our lifetime have structured these areas, shaping our visual perception, including aesthetic preferences. We had seen only a small sample of billions of pictures and art objects that AI systems are trained on. These systems do not blend bits and pieces of images found in training databases. We are aware only of the final stage of processing visual data by the brain, recognizing shapes, objects, landmarks, or people in our field of vision. Intermediate steps are not useful; therefore, we do not need and do not have conscious access to the results of partial processing. Blending in AI systems is done at the intermediate levels before visual images become recognizable. Imagen system (Google Research Brain Team) creates photorealistic images using diffusion models with deep language understanding[40], starting from the noise. In a series of iterations, the random noise distribution is changed to make it similar to those found in images encoded with similar text labels. Low-resolution blurred images emerge, gradually changing into photorealistic phantasies. We may see the preservation of a general style but not a copy of original images or their parts. Language models may describe images created in this way to generate comments that closely resemble a long stream of conscious impressions and create new images based on these comments.

---

[39] https://dalle3.ai/

Although AI software is not perfect, makes errors and confabulations, and lacks deeper understanding, the progress achieved in the last few years is amazing. Can AI understand humans? Analysis of digital footprints (activity on social networks and commercial sites) and images of human faces allowed AI to create more accurate

---

[40] Saharia, Chitwan, William Chan, Saurabh Saxena, Lala Li, Jay Whang, Emily Denton, and others, 'Photorealistic Text-to-Image Diffusion Models with Deep Language Understanding' arXiv.2205.11487

personality profiles (including political, sexual, and religious preferences) than humans could infer using the same data[41]. AI personality model may be more accurate than our ideas of ourselves. The ancient call to "know thyself" has found a surprising solution: we may learn more about ourselves from AI software than from psychotherapists[42].

Making models of spatial structures, understanding relations between the objects, making abstractions, and inferring the probable evolution of situations aimed at the creation of the "multimodal foundational models" is the next challenge for artificial intelligence. A common belief is that AI needs huge databases to learn, and this is a slow process. Humans can learn quickly because we have internalized a lot of experiences in our lifetime, enabling anticipation, association, and understanding based on our experiential foundations. AI models may also learn quickly if we start from a large "foundational model" instead of starting from zero. Embedding human experience in such models requires more than texts and images. Understanding movement, intentions expressed in our interactions, emotions, and motivations requires new databases for training AI systems. The availability of such systems should allow for implementing more ambitious goals of objective hermeneutics – reconstruction of the latent meaning and hidden intentions, a model of the deeper understanding of goals, and observed behavior. At this stage, experts are still needed to train AI systems. Linguists and philosophers may create descriptions of the meaning of metaphors, abstract associations, reasons, and motivations that lead to specific actions.

---

[41] Sandra C Matz, Ruth E Appel, and Michal Kosinski, 'Privacy in the Age of Psychological Targeting', *Current Opinion in Psychology* 31 (2020), 116–21.
[42] Włodzisław Duch, 2012, ibid.

Brain research and AI models clearly show that meaning does not occur only in its symbolic form, and text analysis is insufficient to understand the meaning[43]. Thinking is much more than symbol manipulation. For decades AI researchers have been using symbolic knowledge representations. With the help of computer science students, we have analyzed databases and all available text descriptions of a few hundred dog breeds, trying to find a minimum number of questions that could quickly identify a dog's breed. We have failed to find good rules that associate verbal descriptions with the name of the dog's breed. Applications based on the analysis of silhouettes or photographs, such as Dog Scanner (available on Google Play), work perfectly well. The same applies to other animals, plants, and general object recognition. Even in the simple cases of object recognition, pure text description has strong limitations. Progress in artificial intelligence could not have been achieved using symbolic knowledge representation. Attempts to base machine translation on classical linguistic rules and grammar also have failed. Excellent results are reached by training large NLP neural models on natural texts. They can discover and internalize rules, exceptions, and nuances of natural language in distributed systems but require billions of parameters. Only big brains are capable of using language.

### Large Language Models.

Early attempts to create artificial intelligence systems (sometimes called Good Old-Fashioned AI, abbreviated to GOFAI) were based on knowledge conceptualization, using knowledge engineering techniques to store knowledge discovered by humans in large databases. The hope was that millions of facts stored in some form suitable for machine manipulation should enable intelligent information processing. CyC Corporation has created the largest system of this kind, with millions of concepts, assertions, and 30 million logical rules[44]. The meaning of concepts had to be captured in large, diverse contexts that could not be stored in rules, frames, semantic networks, or other knowledge representation techniques. In the XXI century, the development of powerful computer systems and new machine learning techniques, such as the Generative Pretrained Transformers (GPT), and Generative Adversarial Networks (GAN), allowed the training of huge neural networks containing billions of parameters. Such networks, called Large Language Models, or LLMs, became available to the general public at the end of 2022. GPT-3 by OpenAI was the first example of a system with 175 billion parameters, followed within a few months by ten times larger systems from Google, Microsoft, Samsung, Amazon, Baidu, and other technological giants.

---

[43] Oevermann, U., Konau, E., & Krambeck, J. (1987). Cahp. 21. *Structures of Meaning and Objective Hermeneutics*. In Modern German Sociology (pp. 436-448). Columbia University Press.

Although LLM neural networks are only loosely inspired by the human brain, they can be trained on huge text and image data, internalizing thousands of contexts in which a given concept is used. Training is based on masking parts of the sentences and guessing missing words. It may make an impression that this is how these systems work. Our brains are also embodied predictive machines that develop minds in social environments[45]. Our working memory holds only a few pieces of information. For active working memory to relate the information it contains to the knowledge we have acquired over our lifetime, it must be associated with long-term memory that provides relevant context. In the middle of 2023, LLMs could hold conversations in about 4000 languages (and over 1000 in a spoken language). To provide meaningful answers to humans, their "working memory" has to be activated, priming contextual knowledge contained in LLMs, specifying the role it should assume in answering our questions. In multi-agent conversation framework (called AutoGen framework) several agents, or "personalities", are invoked. Activated in this way, language models are capable of a deeper understanding of questions, generating answers that show ingenuity and surprising emergent properties[46].

One of the most shocking experiments illustrating LLMs capabilities has been performed by Eric Schwitzgebel, David Schwitzgebel, and Anna Strasser. In the article "Creating a Large Language Model of a Philosopher,"[47] they ask, "Can large language models be trained to produce philosophical texts that are difficult to distinguish from texts produced by human philosophers?" To answer this question, they have asked Daniel Dennett, one of the leading philosophers of mind, to write his comments on ten philosophical questions. GPT-3 has general knowledge but had to be trained with Dennett's papers and writings to learn his thinking style. LLMs are stochastic systems that will react to the same questions differently every time they are asked. Our answers also depend on the previous history of our mental states. Four GPT-3 responses for each question were collected and evaluated by 425 participants in this experiment, who tried to distinguish Dennett's answer from ChatGPT output. Even 25 experts on Dennett's work succeeded only 51% of the time. People with no background in philosophy were near chance (20%) trying to distinguish GPT -3's responses from those of a real human philosopher. Although experiments with much more advanced GPT-4 systems have not yet been performed, one can expect more sophisticated answers. We can expect AI systems that will read all philosophical works, attempt to model knowledge of famous philosophers, and attempt to answer questions they never thought about[48].

One objection to the use of LLM systems is based on their confabulations. Human creativity is based on imagery and confabulations, but we filter such ideas as not realistic (except when writing science fiction stories). BVSR, one of the most influential theories of creativity, created by D.T. Campbell[49], is based on blind variation, linking and blending different concepts, followed by selective retention or filtering the most interesting combinations[50]. All LLM models confabulate (this is controlled by a parameter called "temperature" or neural noise) but can not filter confabulations that are not interesting or not realistic. However, this is changing very quickly.

How many original thoughts can humans create that LLMs will not be able to infer if such models will be trained on what we have learned? The hope that embodiment may help us to gain some advantages over computers is misconstrued. Emergence in GPT-4 of the theory of mind, or empathy in contact with patients[51], shows a glimpse of what is coming. ChatGPT has outperformed humans in tests based on the Levels of Emotional Awareness Scale[52], reaching 9.7 points out of 10. A Multimodal Large Language Model (MLLM) can answer questions about the observable world by analyzing images, videos, or sounds, planning robotic actions, and showing embodied understanding in environments with complex dynamics. Google has already shown Palm-E model with

---

562 billion parameters controlling robots using internal neural representations of signals[53]. Using visual behavior modeling, imitation-based learning, with signals from internal sensors, will endow the linguistic concepts with deep, embodied meaning, solving the "symbol grounding problem."[54] Such models can identify and describe emotions from behavioral observations, helping both psychiatrists and patients to understand their emotions. Unique properties of the human mind, such as intuition, insight, imagination, and creativity, can also be implemented in computer software[55] [56].

### Brains and human cognitive limitations.

Perhaps we are reaching the limits of traditional approaches to science. We have started to investigate complex interactions between physical, mental, and cultural worlds.[57] However, to reach a deeper understanding may require the application of tools to discover correlations of many factors and reason in much more complex domains than we are able to. Our cognitive limitations will not allow us to synthesize knowledge from all relevant sources. The speed of learning[58] is influenced by many processes that operate at different time and spatial scales, from microseconds to years, and from molecular to the whole body.

---

[53] Driess, D., Xia, F., Sajjadi, M. S. M., Lynch, C., Chowdhery, A., Ichter, B., Wahid, A., Tompson, J., Vuong, Q., Yu, T., Huang, W., Chebotar, Y., Sermanet, P., Duckworth, D., Levine, S., Vanhoucke, V., Hausman, K., Toussaint, M., Greff, K., … Florence, P. (2023). PaLM-E: An Embodied Multimodal Language Model (arXiv:2303.03378)

[54] Chen, B., Vondrick, C., & Lipson, H. (2021). Visual behavior modelling for robotic theory of mind. Scientific Reports, 11(1), Article 1.

[55] Duch, W. (2007). Intuition, Insight, Imagination and Creativity. IEEE Computational Intelligence Magazine, 2(3), 40–52

[56] Pilichowski, M., & Duch, W. (2013). BrainGene: Computational creativity algorithm that invents novel interesting names. 2013 IEEE Symposium on Computational Intelligence for Human-like Intelligence, IEEE Press, 92–99.

[57] Sapolsky, ibid, chap. 9

[58] Duch, Włodzisław. (2013) "Brains and Education: Towards Neurocognitive Phenomics. In: "Learning while we are connected", Vol. 3, Eds. N. Reynolds, M. Webb, M.M. Sysło, V. Dagiene. pp. 12-23.

Symbolic explanations may be good for relatively simple processes investigated by natural sciences. Language and culture may require models of high complexity that cannot be accurately described in a simplified way. Any attempts to do so will most likely end in confabulations. Our thinking and behavior cannot be fully described in a verbal, symbolic way. What are the limits of the symbolic description of mental states? Is "true interpretation" possible if our mental states and our behavior result from the continuous stream of brain activation patterns? We see only the peaks of this activity in our mental world and never step into the same river. Language has a finite number of symbols, and each concept corresponds to many brain states giving it a slightly different meaning, depending on the context[59]. Deeper mechanisms that influence our decisions are hidden from our introspection[60]. Phenomenology has failed to uncover them, and philosophers started to doubt whether this is possible at all. Eric Schwitzgabel argued convincingly that our beliefs about what we feel and what we are conscious of may be mistaken[61]. Our idea about ourselves is just that, an idea, a model. Alexithymia is the inability to identify and describe emotions and feelings experienced by oneself. Our ability to describe other mental states, not only emotions, is also limited. Psychoanalysis tried to discover the roots of psychological problems in early childhood experiences, but now psychiatrists regard such explanations as confabulations. We understand some mechanisms behind psychotherapy in terms of neuroplasticity and emotional arousal[62].

---

[59] M. J Spivey, *The Continuity of Mind.* (New York: Oxford University Press., 2007).
[60] Duch Włodzisław. (2011) Free Will and the Brain: Are we automata? In: 3rd International Forum on Ethics and Humanism in European Science, Environment and Culture, Ed. M.Jaskuła, B.Buszewski, A. Sękowski and Z. Zagórski, Societas Humboldtiana Polonorum, 2011, pp. 155-170.
[61] Schwitzgebel, Eric. Perplexities of Consciousness. Cambridge, Mass: A Bradford Book, 2011.
[62] Cozolino, Louis. *Why Therapy Works: Using Our Minds to Change Our Brains*. New York: W. W. Norton & Company, 2015.

We have deceived ourselves into thinking that our mental processes are essentially independent of the physical world, but countless experiments show that this is not true. Since 1950, this was the main assumption of cognitivism: ignore the body and the brain; it is just hardware, focus on mental representations. The brain was compared to computer hardware, while the mind was seen as the information processing software that can run on different hardware. These ideas dominated the philosophy of mind in the second half of the XX century. Various computer programs called cognitive architectures were constructed[63]. Large-scale efforts to build ontologies, such as the CyC ontology[64], were started in the mid-1980s and are continued to this day. Knowledge bases containing millions of concepts and tens of millions of assertions and rules were manually created. They were used to reason in expert systems but were not that useful in natural language processing tasks (translation, summarization, dialog systems). They were also rather useless in the image and video analysis or control of robot movements. This exercise in practical ontology and epistemology showed the power and limitations of cognitivism. Ontology, epistemology, and ethics are the three main pillars of philosophy. For millennia, philosophy was a matter of speculation, but recent developments grounded ontology and epistemology in knowledge engineering. With the introduction of autonomous intelligent systems ethics became also a very important practical subject.

Embodied cognitive science has challenged the usefulness of purely symbolic representations and facilitated the move towards distributed neural patterns of activity that implement associative memory, processing of perception, and various mental states. Studies of embodiment became an influential interdisciplinary field of research that is focused on the role of the body in the construction of mental representations[65]. It has been developed further into the 4E cognition theories: embodied, embedded, enactive, and extended[66]. This theory is biologically grounded, viewing cognition as the ability to coordinate and control action in a dynamic environment. Advanced robotics requires a holistic approach that considers the mind and body, or hardware and software, as inseparable. The complexity of the human brain – almost 100 billion neurons and over 100 trillion synapses – did not allow for building robotic models inspired by the 4E theories. However, with the advent of large multimodal models, this is changing, and finer approximations of brain processes result in a more faithful imitation of human-level cognition.

---

[63] Duch, Wlodzislaw, Richard Jayadi Oentaryo, and Michel Pasquier. 2008. "Cognitive Architectures: Where Do We Go from Here?" In Frontiers in Artificial Intelligence and Applications, 171:122–36.
[64] CyC web page contains all details, https://cyc.com

United Nations 2030 Agenda for Sustainable Development has identified several Grand Challenges – Global Health, Sustainable Cities, Energy, Climate, Biodiversity, Cultural Understanding, Human Well-being, and Justice and Equality. The last three challenges are clearly in the domain of humanities. Without artificial intelligence we cannot solve any of these problems. How the change in our perception of the world is expressed in the literature and media? Information overload, multitasking, constant stimulation with news about strange events around the globe, and thousands of images that we see every day have transformed cultural reception and the ways we perceive the world. In the past, literature and arts helped us to broaden our individual perspectives. Semir Zeki has analyzed how brain machinery, individual variability, and the capacity to synthesize available knowledge enabled the development of culture, aesthetic theories, and Platonic ideals[67]. Variability, the driving force of evolution, is also the basis of subcultures, isolating people. The ability to create abstraction and formulate ideals is the uniting force, but it may lead to disappointment. Zeki has called his book "Splendours and Miseries of the Brain." Robert Sapolsky gives a few examples of complex relations between genetics, epigenetics, environment, social structures, and individual personality profiles[68]. We now have tools to analyze relations between many types of such data.

---

[65] Damasio, Antonio. *The Feeling of What Happens: Body and Emotion in the Making of Consciousness.* San Diego, CA: Mariner Books, 2000; Johnson, Mark. The Body in the Mind: The Bodily Basis of Meaning, Imagination, and Reason. University of Chicago Press, 2013.
[66] Barrett, L. (2018). The Evolution of Cognition: A 4E Perspective, pp. 719–734. In A. Newen, L. De Bruin, & S. Gallagher (Eds.), The Oxford Handbook of 4E Cognition. Oxford University Press.

Methods used in humanities that rely on critical interpretations are subjective and open to cognitive biases. Daniel Kahneman received the Nobel prize in 2002 for his work on heuristics and biases in decision-making. His popular book "Thinking fast and slow"[69] was followed by dozens of books describing hundreds of cognitive biases (Wikipedia lists over 240 types of biases). We are not training our students to avoid and spot such biases. Are we sure that research in humanities is not affected by cognitive biases? The history of medicine is full of terrible mistakes that lasted for centuries: the use of mercury, bloodletting for almost every ailment, humoral theory, animal magnetism, and homeopathy all gave people the illusion of understanding. Romantic medicine, influenced by Keats and Goethe, survived until the middle of the 19th century. Goethe wrote in Faust: 'Similia Similibus applies to all disorders, expressing his support for the homeopathic idea that a disease may be cured by something that can cause similar symptoms. To this day, homeopathy is based on this superstition. A lot of speculative theories that were not supported by any evidence survived from ancient times, and new pseudo-scientific therapies are created each year. A few random correlations and superficial similarities are given as explanations. Rhinoceros horns are not aphrodisiacs, mercury (called in the past quicksilver) is not the elixir of life, and human character is not related to the shape of our skulls or faces, as claimed by phrenology, psychognomy, physiognomy. Psychology is also full of constructs derived from common-sense understanding of mental processes and behavior. In the last decades, neuropsychology has distinguished more than 10 types of memory and a similar number of attention types that depend on specific brain processes[70]. Psychological constructs are slowly being aligned with the brain processes. Medicine and some branches of psychology are now firmly linked to physical sciences. In the case of humanities, this is much more difficult, as we deal here with higher-level mental processes that cannot be investigated in simple experimental situations. Semir Zeki, Mark Johnson, Eric Schwitzgebel, and many others showed that analysis of our inner experience might go beyond folk psychology concepts.

---

[67] Semir Zeki. 1999. "Splendours and Miseries of the Brain." Philosophical Transactions of the Royal Society of London. Series B: Biological Sciences 354 (1392): 2053–65, and Semir Zeki, Splendors and Miseries of the Brain: Love, Creativity, and the Quest for Human Happiness (John Wiley & Sons, 2011).
[68] Robert Sapolsky, ibid, chap. 9
[69] Daniel Kahneman, *Thinking, Fast and Slow* (Farrar, Straus and Giroux, 2011).

Disillusion with mainstream knowledge (requiring serious effort to understand) ends in believing in superstitions (easy to understand) that are spread in a charismatic way. There is a strong competition to own our minds, fill our heads with cultural narratives, and impose new structures on our conceptual network. Which stories that we hear every day sink into our heads and control our thoughts? Instead of evaluating arguments, people tend to listen to opinions and speculations, following their emotions. To survive, people (like all social animals) had to conform to local culture. In the age of communication technologies accessible to billions of people, a multitude of subcultures propagate false beliefs and conspiracy theories. Without training in critical thinking, people tend to believe what they read or are told. Verification is of great importance in many branches of humanities, but without specialized tools, it is rarely possible. Loss of interest in humanities may be related to the information overflow. Constant distractions, immediate access to information, multitasking, and jumping from one topic to another on the Internet and social media leave little time for exploration of interesting subjects. Instead of reading books, children learn to read abbreviated versions that allow them to answer examination questions. Scientists judge papers by their abstracts or do a quick and shallow reading.

---

[70] Duch, W. (2018). Kurt Lewin, psychological constructs and sources of brain cognitive activity. Polish Psychological Forum 23(1), 7–21

Scientific discoveries are translated into applications in technology. We cannot doubt the existence of electromagnetic waves. There is little consensus on the meaning of concepts used in philosophy, political sciences, interpretation of ancient texts, poetry, art, or music critique.

## Current situation of humanities

In "The Challenges to the Humanities" Patricia Meyer Spacks, a famous American literary scholar, wrote: "Ever since I can remember—and I go back a way now—humanists have been declaring a crisis."[71] After more than 20 years, these words are still true. She has described three challenges: funding, conflict, and communication. Public support of culture and humanities research in the USA is minimal. Culture has to earn money and relies on public donations. For example, only 3% of the budgets of 138 symphonic orchestras in the USA came from government sources (League of American Orchestras, 2020)[72]. Conflict refers to extreme fragmentation and lack of agreement in many fields of humanities. Communication of the value of humanities is quite poor, and public opinion hears more about political accusations (usually left-wing) than interesting developments in moral foundation theory[73].

---

[71] Patricia Meyer Spacks, 'The Challenges to the Humanities', *American Academy of Arts & Sciences* Winter 2001 Bulletin edition (Cambridge, MA, 2001), <https://www.amacad.org/news/challenges-humanities> [accessed 30 August 2022].
[72] https://americanorchestras.org/wp-content/uploads/2020/11/Orchestras-at-a-Glance-2020.pdf

A recent report for the Higher Education Policy Institute (HEPI) in the UK is not optimistic, focusing on funding and encouraging high school students to take humanistic subjects[74]. Between 1962 and 2010, the proportion of UK students studying Humanities subjects fell from around 28 to around 9% of all students. Teaching foreign languages may soon be done on a much smaller scale. Several British universities have canceled many humanities courses in classics, philosophy, literature, and languages. Recruitment for modern language courses is very low. Online translation services are doing a very good job, and smartphone applications translate speech in an instant, so why would anyone need to learn a foreign language?

The situation in the European Union is a bit better, thanks to many programs under the "Creative Europe" and "Culture, creativity and inclusive society" heading. Most research funds for social sciences and humanities in the EU framework programs are integrating them with big challenges, like an assessment of the societal impact of natural, physical, health sciences, or technology. These programs are directed mainly at social scientists, but some projects can also benefit the humanities. Among many other initiatives, they include the preservation of rare languages, cultural heritage, traditional crafts, arts, and music, but also the impact of games on society, their risks, cultural value, and innovation potential. The New European Bauhaus (NEB) initiative is aimed at designing public and private spaces that should promote well-being and a sense of belonging involving science, technology, art, and culture.

---

[73] https://moralfoundations.org
[74] Gabriel Roberts, The Humanities in Modern Britain: Challenges and Opportunities. HEPI Report 141. https://www.hepi.ac.uk/wp-content/uploads/2021/09/The-Humanities-in-Modern-Britain-Challenges-and-Opportunities.pdf

All these efforts are unlikely to improve the situation of humanities significantly. There is a high discrepancy between the salaries of graduates among STEM majors and humanities. Fortunately, the latest developments in digital humanities may help to break the barriers between many academic fields.

### The times they are a-changin'.

We are experiencing a tsunami of changes that started with personal computers and the Internet, the creation of freely accessible information sources, repositories of research papers, encyclopedias, and digital museums. We now have access to the digitized versions of almost all texts, images, and videos ever created. Some people understood the power of digitization quite early. In 1949 Father Roberto Busa, a Jesuit priest, persuaded the founder of IBM, Thomas J. Watson, to sponsor the Index Thomisticus. The creation of a digitized version of Thomas Aquinas's works and the development of a tool for performing text searches lasted about 30 years, finally leading to a web-based version published in 2005. Such projects can now be done quite quickly. Taking advantage of the opportunities provided by digital humanities and new artificial intelligence tools gives us a great chance to overcome our cognitive limitations in projects of such a large scale.

Although not everyone will embrace these new opportunities, the number of scholars using computational methods in humanities to collect, analyze, and visualize data is growing. The Social Sciences and Humanities Open Cloud (SSHOC) project is a European endeavor that brings together a cluster of research infrastructures serving the social sciences and humanities to build services related to the European Open Science Cloud[75] . Digital tools are developed for the protection, identification, and traceability of cultural goods. CLARIN (Common Language and Technology Infrastructure)[76] and DARIAH (Digital Research Infrastructure for the Arts and Humanities)[77] are two big consortia providing a vast number of freely accessible online collections of cultural heritage data, serving the humanities and social sciences. The DARIAH consortium, established in 2014, has been joined by most European countries and the USA. It has working groups on many special subjects, such as the Theatralia group, covering all kinds of digital support approaches in the performing arts, including scenography, costume design, setup of sounds and projections, and trying to capture intangible heritage aspects. Working group on Music and Artificial Intelligence [78] brings together academic researchers and various institutions, including commercial companies, investigating applications of intelligent systems to automatic composition, music generation, semantic annotation, music teaching, analysis of emotions, and the integration of music with different branches of art.

---

[75] https://sshopencloud.eu

In 2020, the European Union started to support four large consortia working on the application of AI to real-life problems. Two of these consortia are especially interesting for humanities. AI4Media [79] is a very broad consortium supporting the multidisciplinary community that works on all aspects of applications of AI to media: support for content creation (news, graphics, video), presentation and human co-creation, the influence of media on society, the economic and political impact of AI technologies, spreading disinformation, threats to democracy, and other concerns. This consortium has established The European AI Media Observatory[80], a knowledge platform to monitor research on AI in media. A subgroup AI for Social Sciences and Humanities[81] provides tools for the identification of patterns in aggregated, multimodal collections. These tools should allow for macro-level analysis of political bias in media, opening novel ways to do research in political sciences. HumanE-AI-net, the European Network of Human-centered Artificial Intelligence,[82] is another large consortium with 53 partners (as of 6/2023) concerned with the development of robust, trustworthy AI systems that enhance human capabilities, understand and empower people acting in real-world complex situations while respecting their autonomy. For example, the AI for Human Memory project, aimed at understanding, supporting, and improving human memory, brings together experts from psychology, Human-Computer Interaction (HCI), and computer science to discuss memory-related technology from a human-centered AI perspective.

---

[76] https://www.clarin.eu
[77] https://www.dariah.eu
[78] https://ai-music.org
[79] https://www.ai4media.eu
[80] https://www.ai4media.eu/observatory/

AI4Europe [83] wants to create a sustainable, integrative digital platform and experimentation environment for the European AI research ecosystem to maximize its academic, social, and industrial impact. It should help to enhance human capabilities to understand and interact with humans in complex social settings, empowering them to achieve more. Such systems should respect law and ethics "by design." These objectives will not be easy to achieve but will increase demands for humanities graduates.

AIDA, The Artificial Intelligence Doctoral Academy, offers training in general AI and machine learning techniques and specialized courses from AI4Media, HumaneAINet, and other EU consortia. The annual report "Journalism, media & technology in 2023 – Trends & Predictions" from Reuters Institute[84] is focused mostly on AI. Many universities across the world have established their own digital humanities centers. Over the last decades, vast numbers of digital collections and tools have been created for digital humanity research, opening many opportunities for interdisciplinary research and showing the path toward humanity's revival.

---

[81] https://www.ai4media.eu/uc4-ai-for-social-sciences-and-humanities/
[82] https://www.humane-ai.eu
[83] https://www.ai4europe.eu
[84] https://reutersinstitute.politics.ox.ac.uk/journalism-media-and-technology-trends-and-predictions-2023

All this was just the beginning. In the last few years, artificial intelligence has given us new tools that can be used to analyze all kinds of data. AI started to profoundly change our lives, expanding human experience beyond access to information. In 2019, Stanford University created the Institute for Human-Centered AI[85], trying to understand what impact new technologies will have on society. Many AI projects are based on this human-centered approach to AI. Large language models are trained using reinforcement learning with human feedback (RLHF)[86] to align their answers with human preferences. Such training cannot be done without the participation of psychologists, sociologists, lawyers, and philosophers.

Artificial intelligence is transforming the world as we know it, creating a future where AI will impact humans in ways we cannot fully imagine today. From academia to industry and government, everyone is seeking to understand the impact of AI on our future. Can it be just a fad, techno-enthusiasm that will finally vanish? Many people could not believe that personal computers, and later the Internet, will significantly change the way we live. Recognizing factors that influence people's perceptions, beliefs, values, and traditions and contribute to their well-being is a great challenge. Understanding subcultures, their values, ambitions, and motivations requires analysis of the economy, history, culture, attitudes related to gender issues, aging of societies, manifestations of racism, xenophobia, and conspiracy beliefs. Artificial intelligence can be of great help in this endeavor. Unfortunately AI can also be used to enslave people. Understanding human psychology, recognizing personality features, and ability to act autonomously in the Internet will make it easy to manipulate people. An overview of Catastrophic AI risks is presented at the Center for the AI safety page[87].

---

[85] https://hai.stanford.edu
[86] https://en.wikipedia.org/wiki/Reinforcement_learning_from_human_feedback

It will be criticized as shallow, techno-enthusiastic, and misguided by the defenders of traditional humanities, pointing to errors and confabulations of current large language models. However, all these technologies are transitional; errors will soon be corrected, and the incredibly fast progress will only accelerate.

We should look at AI and digital humanities as a great opportunity for the revival of all humanity subjects, increasing the interest of bright students, and a chance to create new, ambitious, large-scale projects that can lead us beyond our cognitive limitations.

---

Włodzisław Duch
Katedra Informatyki Stosowanej i Laboratorium Nuerokognitywne
Uniwersytet Mikołaja Kopernika
https://orcid.org/0000-0001-7882-4729

# Sztuczna inteligencja i granice nauk humanistycznych.

Złożoność kultur we współczesnym świecie przekracza obecnie możliwości ludzkiego pojmowania. Nauki kognitywne poddają w wątpliwość tradycyjne wyjaśnienia oparte na modelach mentalnych. Tradycyjne dyscypliny humanistyki mogą tracić na znaczeniu, ale pojawiają się nowe, interdyscyplinarne gałęzie humanistyki. Humanistyka musi dostosować się do epoki cyfrowej. Błyskawiczny dostęp do informacji zostanie wkrótce zastąpiony błyskawicznym dostępem do wiedzy. Sztuczna inteligencja zmieni humanistykę w sposób radykalny, od sztuki po nauki polityczne i filozofię. Zrozumienie ograniczeń poznawczych człowieka i możliwości, jakie otwiera rozwój sztucznej inteligencji oraz interdyscyplinarnych badań niezbędnych do podjęcia globalnych wyzwań, jest kluczem do rewitalizacji humanistyki.

Wlodzislaw Duch works at the Nicolaus Copernicus University in Toruń, Poland. His PhD/DSc was in theoretical physics, but since 1990 he worked on artificial intelligence and cognitive science in the USA, Singapore, Japan, Germany and other countries. He is/was on the editorial board of 17 international journals; served as the President of the [European Neural Networks Society](), is a fellow of the International Neural Network Society, a member of 3 Polish Academy of Science committees. Published about 370 peer-reviewed scientific papers, has written or co-authored 6 books and co-edited 21 books. Type "Wlodzisław Duch" in the Internet to find his full CV.

Włodzisław Duch pracuje na Uniwersytecie Mikołaja Kopernika w Toruniu. Doktorat/habilitację zrobił z fizyki teoretycznej, ale od 1990 roku pracował nad sztuczną inteligencją i kognitywistyką w USA, Singapurze, Japonii, Niemczech i innych krajach. Jest/był członkiem rad redakcyjnych 17 międzynarodowych czasopism; pełnił funkcję prezydenta European Neural Networks Society, International Neural Network Society Fellow, jest członkiem 3 komitetów Polskiej Akademii Nauk. Opublikował około 370 recenzowanych prac naukowych, jest autorem lub współautorem 6 książek i współredaktorem 21 książek. Aby znaleźć jego pełne CV wystarczy wpisać jego imię i nazwisko.